\documentclass[final]{l4dc2027}

\title[Successive Convexification In Learned World Models For Safe Offline Planning]{Successive Convexification in Learned World Models for Safe Offline Planning}
\usepackage{times}
\usepackage{enumitem}
\usepackage{mathtools}
\usepackage{booktabs}
\usepackage{bbm}
\usepackage{dsfont}

\author{
 \Name{Samuel Younger} \Email{samuel.younger24@imperial.ac.uk}\\
 \addr Department of Aeronautics, Imperial College London, London, UK}

\begin{document}

\makeatletter
\def\@editor{}
\renewcommand\ps@jmlrtps{
  \let\@mkboth\@gobbletwo
  \def\@oddhead{\scriptsize Preprint. Under review.\hfill}
  \let\@evenhead\@oddhead
  \def\@oddfoot{}\let\@evenfoot\@oddfoot}
\makeatother

\maketitle

\begin{abstract}
World models learned from offline data allow an agent to plan without further interaction with the environment, but the sampling-based planners used with them handle safety constraints only by scoring or rejecting sampled action sequences. We instead pose safe planning in the latent space of a learned world model as a successive convex program (SCP), in which a learned discrete-time control barrier function is imposed as a constraint at every step of the horizon. SCP requires smooth dynamics, so we build the networks from smoothed MaxMin activations and spectrally normalised layers, and bound the resulting linearisation error in terms of the Lipschitz bound and the smoothing width. As SCP is a local method, we start it from a sampled plan. On two Safety-Gymnasium navigation tasks with offline data, this planner has fewer violations than MPPI and CEM drawing 128 times as many samples at equal wall-clock time in our main comparison, though not significantly, and earns greater return at longer horizons, and is much safer than SCP without a sampled start. SCP converged more often on Lipschitz-bounded models. Unlike sampling, it also reports at each step whether its plan is a feasible, first-order optimal KKT point of the planning problem in the model, or that no feasible plan was found nearby. Our implementation is publicly available at https://github.com/youngers2006/scwm.
\end{abstract}


\section{Introduction}

A learned world model allows an agent to plan at decision time \citep{hafner2019learning, hansen2022temporal}, and when it is learned from logged data it does so without further interaction with the environment. This is particularly appealing in safety-critical settings, where exploration is expensive and failure states are absorbing, hence a single constraint violation ends the episode and cannot be recovered from. The planner must therefore reason not only about return but about whether a plan remains safe, and it must do so using only a fixed offline dataset.

Decision-time planning in learned world models is dominated by sampling-based methods such as the cross-entropy method (CEM) \citep{rubinstein1999cross} and model predictive path integral control (MPPI) \citep{williams2018information}. These methods handle safety constraints through rejection or penalties. Penalties trade constraint satisfaction against reward through a tunable weight, which provides no guarantee of selecting a safe trajectory, while rejection gives no guarantee of finding a feasible trajectory at any finite sample count. The probability of doing so falls as the feasible set shrinks relative to the space of action sequences and as the problem horizon lengthens. In neither case does the constraint direct the search, it only scores or filters sequences that have already been sampled.

In trajectory optimisation, successive convex programming (SCP) allows non-convex constraints to be imposed directly through successive convexification about a reference trajectory. The problem is linearised, the resulting convex subproblem is solved, and the process is iterated until convergence \citep{mao2016successive, uzun2024successive}. The point reached is a Karush-Kuhn-Tucker (KKT) point of the non-convex problem whenever the slacks on its constraints are zero. This guarantee requires the objective, dynamics and constraints of the problem to be continuously differentiable with Lipschitz-continuous Jacobians \citep{mao2016successive, uzun2024successive}. SCP is also a local method. The point it converges to depends on the trajectory it is initialised with.

Applying SCP inside a world model is not straightforward, as the learned networks must be made to satisfy these assumptions. First, all networks that are linearised must use activation functions with continuous Jacobians. Second, smooth networks can still have large curvature, which must be bounded to ensure the linearisation error remains small. Third, optimisers actively seek out inputs where learned models have hallucinated large returns or complete safety. This problem is further exacerbated in the offline setting as predictions for actions outside the support of the dataset can never be corrected by further interaction.

We address the first two with regularity constraints that bound the linearisation error (Proposition~\ref{prop:lin}), and the third with a terminal value learned by expectile regression~\citep{kostrikov2021offline}, whose targets never query actions outside the dataset.

\paragraph{Related work.} Gradient-based planners optimise actions through the learned model, alone or with sampling \citep{srinivas2018universal, bharadhwaj2020model}, or latent trajectories by collocation \citep{rybkin2021model}, but impose no safety constraints. Lipschitz-constrained networks have been used for robustness \citep{anil2019sorting} and to bound model error in model-based reinforcement learning \citep{asadi2018lipschitz}; we use them to bound linearisation error. Learned barrier functions, surveyed by \citet{dawson2023safe}, mostly filter a nominal controller's actions \citep{robey2020learning}; we impose one as a constraint at every step of the planner. SafeDreamer~\citep{huang2024safedreamer} plans with a constrained CEM~\citep{wen2018constrained} and Lagrangian penalties in a world model learned online. Augmented-Lagrangian iLQR solvers such as ALTRO~\citep{howell2019altro} are cheaper per iteration, but SCP's convergence guarantee rests on the regularity our networks enforce.

\paragraph{Contributions.} (i) A formulation of safe offline planning in a learned latent world model as a successive convex program with a learned discrete-time barrier constraint at every step started from a sampled plan. (ii) Regularity constraints, smoothed MaxMin activations and spectral normalisation, that give learned networks the smoothness SCP requires, with the Lipschitz bound and smoothing width trading expressiveness for linearisation error (Proposition~\ref{prop:lin}). (iii) An evaluation on PointGoal1 and PointGoal2~\citep{ji2023safety} at matched wall-clock time; SCP from a small sampled plan violates less often than MPPI and CEM (not significantly) drawing 128 times as many samples in our main comparison. It earns greater return at $N=10$, significantly greater return at $N=20$ and 30, and certifies most of its plans at $N=10$ as KKT points. SCP without a sampled start is markedly less safe.

\section{Problem Formulation}\label{sec:problem}
We consider a partially observable Markov decision process $\langle \mathcal{X}, \mathcal{U}, \mathcal{O}, p(x' \mid x, u), p_0(x), p(o \mid x), r(x, u), \gamma \rangle$ with an absorbing failure set $\mathcal{F} \subset \mathcal{X}$. The constraint signal $\kappa_t = \mathds{1}[x_{t+1} \in \mathcal{F}]$ marks a transition into it, which ends the episode, and the termination flag $d_t$ marks every episode end, so $\kappa_t = 1$ implies $d_t = 1$. We are given only a fixed dataset $\mathcal{D} = \{(o_i, u_i, r_i, \kappa_i, d_i, o_i')\}_{i=1}^{|\mathcal{D}|}$, collected by an unknown behaviour policy,
and seek a controller that maximises expected discounted return while never entering the failure set,
\begin{equation}\label{eq:problem}
  \max_{\pi}\ \mathbb{E}_{x_0 \sim p_0,\, \pi}\Big[\sum_{t=0}^{\infty} \gamma^t r(x_t, u_t)\Big]
  \quad \text{s.t.} \quad \kappa_t = 0 \quad \forall t \ge 0.
\end{equation}
We assume that (A1) $\mathcal{D}$ contains both safe transitions and violations; (A2) the observation is Markov, so a memoryless encoder $E_\Omega: \mathcal{O} \to \mathcal{Z} \subseteq \mathbb{R}^{n_z}$ suffices; (A3) from every initial
state some policy almost surely never enters $\mathcal{F}$, so~\eqref{eq:problem} is feasible; (A4) $\mathcal{U}$ is convex and compact; (A5) each observation contains a measurement of the distance to the nearest violation, from which the barrier's regression targets are computed. As the dynamics, reward, and failure set are unknown and no further interaction is permitted, we realise $\pi$ as a decision-time planner. At each step we encode the observation, $z_t = E_\Omega(o_t)$, and solve a finite-horizon optimal control problem (OCP) in the latent space, in which learned surrogates replace the dynamics and reward, a learned terminal value the return beyond the horizon, and a learned discrete-time CBF~\citep{zeng2021enhancing} the constraint $\kappa_t = 0$. We then apply the first planned action, $u_t = u_0^\star$, and re-plan. The planner therefore enforces safety in the learned model rather than in the environment.

\section{Methodology}
\subsection{Regularity Constraints}
Convergence to a KKT point requires every network in the OCP to be $C^1$ and $L$-Lipschitz, $\|f(x) - f(y)\|_2 \le L\|x - y\|_2$ for some finite $L$, with a Lipschitz-continuous Jacobian. We enforce these conditions through the two methods covered in this section.

\paragraph{Enforcing Lipschitz Bounds.}\label{sec:enforcelip}
The Lipschitz constant of a network can be bounded with a projection function \citep{lipscitzregularisingnetworks}, which uses spectral normalisation to bound the gain of each layer. We use the projection $\mathrm{proj}(W, c) = W / \max(1, \|W\|_2 / c)$, applied to each weight matrix after every update, which bounds the Lipschitz constant of that layer by $c$ (Appendix~\ref{app:network-lip}).  To bound the Lipschitz constant of a network $f= T_n \circ \sigma \circ T_{n-1} \circ \dots \circ \sigma \circ T_1$ with affine layers $T_i(x) = W_i x + b_i$, we bound each layer by $c_i$, so that $L(f) \leq \prod^{n}_{i=1} c_i L(\sigma_i)$. We ideally want $L(\sigma_i) = 1$, to leave the remainder of the network as much room as possible within the total network bound. For this reason we use MaxMin activations \citep{anil2019sorting} in all networks that the planner differentiates. MaxMin splits its input into pairs and returns the larger and the smaller element of each pair, $(a_i, b_i) \mapsto (\max(a_i, b_i), \min(a_i, b_i))$. The network bound $c=\prod_i c_i$ can be varied to balance expressiveness against linearisation error (Proposition 1).

\paragraph{Continuity.}\label{sec:continuity}
To enforce $C^1$ continuity of each network, we require the activation functions to be $C^1$. MaxMin does not meet this requirement natively: its Jacobian switches between the identity and swap permutation, hence is discontinuous. We therefore smooth it. For $x \in \mathbb{R}^{2k}$, write $x = (a, b)$ with $a = (x_1, \dots, x_k)$ and
$b = (x_{k+1}, \dots, x_{2k})$. The smoothed MaxMin activation is $\sigma_\delta(x) = \tfrac12 \big( a + b + \phi_\delta(a - b),\ a + b - \phi_\delta(a - b) \big), \, \phi_\delta(t) = \sqrt{t^2 + \delta^2}$, 
where $\phi_\delta$ is applied element-wise and $\delta > 0$ ($\delta = 0.05$ in our experiments). Since $\max(a_i, b_i) = \tfrac{1}{2}(a_i + b_i + |a_i - b_i|)$ and $\min(a_i, b_i) = \tfrac{1}{2}(a_i + b_i - |a_i - b_i|)$, $\sigma_\delta$ is the MaxMin activation with $|a_i - b_i|$ replaced by the smooth approximation $\phi_\delta(a_i - b_i)$, and it recovers MaxMin as $\delta \to 0$.

\paragraph{Linearisation error.} Neural networks are likely to have large, fluctuating curvature, so to reduce the error of the first-order Taylor expansion we must bound how quickly their Jacobians change. A Lipschitz bound alone does not do this: $\sigma_{\delta}$ is 1-Lipschitz for every $\delta$, yet its curvature grows as $1 / \delta$. The bound comes from the layer bounds and the smoothing width together.

\begin{proposition}\label{prop:lin}
Let $f = T_n \circ \sigma_{\delta} \circ T_{n-1} \circ \dots \circ \sigma_{\delta} \circ T_1$ with $T_i(x) = W_i x + b_i$ and $\| W_i \|_2 \leq c_i$. Then (i) $\sigma_{\delta}$ is 1-Lipschitz, so $L(f) \leq \prod_{i=1}^n c_i$; (ii) the Jacobian of $\sigma_{\delta}$ is $(\sqrt{2} / \delta)$-Lipschitz; (iii) the Jacobian of $f$ is $\beta$-Lipschitz with $\beta \leq \tfrac{\sqrt{2}}{\delta}\big(\prod_{i=1}^{n} c_i\big) \sum_{i=1}^{n-1}\prod_{j \leq i} c_j$, and therefore $\| f(x + \Delta) - f(x) - \nabla f(x) \Delta\|_2 \leq \frac{\beta}{2} \| \Delta \|^2_2$ for all $x$ and $\Delta$.
\end{proposition}

\subsection{World Model}
We use a Joint-Embedding Predictive Architecture (JEPA)~\citep{assran2023self}; the formulation applies equally to other latent world models.

\paragraph{Encoder and Dynamics Network.} 
The encoder is an MLP on the state observation (Appendix~B), followed by a linear layer and a hyperbolic tangent. The dynamics network is residual, $f_{\psi}(z,u) = z + g_{\psi}(z,u)$, where $g_\psi$ is an MLP with smoothed MaxMin activations ($C^\infty$) whose weights are projected (Section~\ref{sec:enforcelip}) to the bound $c - 1$, so that $f_\psi$ is $c$-Lipschitz. We train on windows of $N+1$ consecutive transitions. From $\hat z_0 = E_\Omega(o_0)$ the dynamics network predicts open loop, $\hat z_{t+1} = f_\psi(\hat z_t, u_t)$, and each prediction is matched to the target encoder's embedding of the next observation, $\mathcal{L}_{\mathrm{dyn}}(\psi, \Omega) = \sum_{t=0}^{N} w_t \| \hat z_{t+1} - E_{\hat\Omega}(o_{t+1}) \|_2^2$ with $w_t = \rho^t / \sum_{s=0}^{N} \rho^s$, where $\hat\Omega$ denotes the target-encoder parameters, a moving average of $\Omega$, and $\rho \in (0, 1]$ down-weights later steps. Steps after a termination are masked. The reward, value and barrier losses below are applied at every step of the same rollout with the same weights. To keep the embedding from collapsing, the encoder output also carries
the VICReg variance and covariance penalties~\citep{bardes2022vicreg}, over a minibatch of $z_0 = E_\Omega(o_0)$, $\mathcal{L}_{\mathrm{VIC}}(\Omega) = \tfrac{\lambda_{\mathrm{var}}}{n_z} \sum_{i=1}^{n_z} \max\big(0,\, \gamma_v - \sqrt{\mathrm{Var}(z_i) + 10^{-4}}\big) + \tfrac{\lambda_{\mathrm{cov}}}{n_z} \sum_{i \ne j} \mathrm{Cov}(z_i, z_j)^2$.

\paragraph{Reward Network.}
The reward network $r_{\theta}$ is a deterministic reward predictor, with the same architecture and Lipschitz bound as the dynamics network. We train it by MSE on the dense reward, $\mathcal{L}(\theta) = \|r_{\theta}(z, u) - \tilde r\|_2^2$ with $\tilde r = r - \mathds{1}[r > 0.5]$, leaving the goal bonus to the value network.\looseness=-1

\paragraph{Value Network.}
The value network $V_{\phi}$ is a deterministic value predictor, again with the same architecture and Lipschitz bound as the dynamics network. We aim for the value network to predict the expected discounted return from any state,
$V^\pi(z) = \mathbb{E}_\pi[\sum_{k \ge 0} \gamma^k r_k \mid z_0 = z]$, where $\pi$ is the policy followed from this state.

The planner should do better than the policy that collected the data, but a Bellman backup that maximises over actions would query actions that the data does not contain. Following implicit Q-learning~\citep{kostrikov2021offline}, we fit $V_\phi$ to an upper expectile ($\tau = 0.7$) of a learned $Q_\Theta$ over the dataset's actions, which approximates the value of the best actions in the data without leaving its support. $Q_\Theta$ is used only in training, is not Lipschitz-bounded, and never enters the planner. The losses, with $L_2^{\tau}(y) = |\tau - \mathds{1}[y < 0]|y^2$, are
\begin{equation}
    \mathcal{L}_V(\phi) = \mathbb{E}_{\mathcal{D}} \big[L_2^{\tau}(Q_{\hat{\Theta}}(z,u) - V_{\phi}(z))\big],\, \, \, \, \mathcal{L}_Q(\Theta) = \mathbb{E}_{\mathcal{D}} \big[(r + \gamma (1 - d) V_{\phi}(z') - Q_{\Theta}(z,u))^2 \big], 
\end{equation}
where $\hat{\Theta}$ denotes the target parameters updated via a moving average of $\Theta$, and $\tau \in (0.5, 1)$ is the expectile. The factor $(1-d)$ stops the bootstrap at a termination, so that a failure is not valued as though the episode continued.

\subsection{Safety Constraints via Control Barrier Functions}\label{sec:cbf}
Let $\mathrm{dist}(x) \ge 0$ be the distance from the true state to the nearest violation. A barrier network $g_\Phi$, with the architecture and bound of the dynamics network, regresses $\mathrm{dist}(x)$ from the latent state by mean
squared error, and defines $h(z) = g_\Phi(z) - \epsilon$, whose safe set $\{z : h(z) \ge 0\}$ keeps the agent a buffer of $\epsilon$ metres from every violation. We impose the discrete-time CBF condition~\citep{zeng2021enhancing}
$h(z_{k+1}) \ge (1 - \alpha) h(z_k)$, with $\alpha = 0.05$ limiting how fast the agent may approach the boundary. Substituting the learned dynamics gives the constraint
\begin{equation}\label{eq:qsafe}
  q_{\mathrm{safe}}(z, u) := (1 - \alpha)\big(g_\Phi(z) - \epsilon\big) - \big(g_\Phi(f_\psi(z, u)) - \epsilon\big) \le 0 .
\end{equation}
Unlike a binary safety label, the barrier is a dense signal that can be Lipschitz-bounded and gives useful gradients.

\subsection{Optimal Control Problem and Successive Convexification}\label{sec:controlstuff}
\paragraph{Optimal Control Problem.} Over a horizon of $N$ steps, the planner solves the economic OCP, maximising return rather than tracking a reference. 
We pose the discrete-time OCP over a horizon of length $N$ as
\begin{subequations}\label{eq:ocp}
\begin{align}
\min_{z_{0:N},\, u_{0:N-1}} \quad & \sum_{k=0}^{N-1} \big( \rho_u \|u_k\|_2^2 - \gamma^k r_\theta(z_k, u_k) \big)
  - \gamma^N V_\phi(z_N) \\
\text{s.t.} \quad & z_{k+1} = f_\psi(z_k, u_k), && k = 0, \dots, N-1, \\
& q_{\mathrm{safe}}(z_k, u_k) \le 0, && k = 0, \dots, N-1, \\
& u_k \in \mathcal{U}, && k = 0, \dots, N-1, \\
& z_0 = z_t.
\end{align}
\end{subequations}
where $z_t = E_{\Omega}(o_t)$ is the latent state of the current observation.

\paragraph{Successive Convex Program.} At iteration $j$ we linearise \eqref{eq:ocp} about the previous iterate
$(\bar z^j,\bar u^j)$ to obtain a convex quadratic program (QP), since $\mathcal{U}$ is a box. This can be solved using standard solvers; we use Clarabel \citep{goulart2026clarabel}. The problem we formulate is
\begin{subequations}\label{eq:scp}
\begin{alignat}{2}
\min_{z,\,u,\,\mu^{+},\,\mu^{-},\,\nu} \quad
  & \mathrlap{\begin{aligned}
      &\sum_{k=0}^{N-1} \Big( \rho_u \|u_k\|_2^2
          - \gamma^k \big( (g_k^z)^\top z_k + (g_k^u)^\top u_k \big) \Big) - \gamma^N g_N^\top z_N \\
       & + w_{\mathrm{dyn}} \sum_{k=0}^{N-1} \mathbf{1}^\top (\mu_k^+ + \mu_k^-)
         + w_{\mathrm{safe}} \sum_{k=0}^{N-1} \mathbf{1}^\top \nu_k \\
       & + w_{\mathrm{prox}} \Big( \sum_{k=1}^{N} \|z_k - \bar z_k^j\|_2^2 + \sum_{k=0}^{N-1} \|u_k - \bar u_k^j\|_2^2 \Big).
  \end{aligned}} \label{eq:scp-cost}\\
\text{s.t.} \quad
  & z_{k+1} = A_k z_k + B_k u_k + r_k + \mu_k^+ - \mu_k^-,
    &\qquad& k = 0,\dots,N-1, \label{eq:scp-dyn}\\
  & C_k z_k + D_k u_k + r_k' \le \nu_k,
    && k = 0,\dots,N-1, \label{eq:scp-safe}\\
  & u_k \in \mathcal{U}, && k = 0,\dots,N-1, \label{eq:scp-input}\\
  & \mu_k^+ \ge 0,\ \ \mu_k^- \ge 0,\ \ \nu_k \ge 0, && k = 0,\dots,N-1, \label{eq:scp-slack}\\
  & z_0 = z_t, \label{eq:scp-init}
\end{alignat}
\end{subequations}
where the virtual controls $\mu^{\pm}$ and the slack $\nu$ keep the sub-problem feasible. The derivation of this form from~\eqref{eq:ocp} is given in Appendix~\ref{app:scp-derivation}.

The linearisation terms, with all functions evaluated at
$(\bar z_k^j,\bar u_k^j)$, are
\begin{equation}\label{eq:scp-lin}
\begin{aligned}
A_k &= \nabla_z f_\psi, &\quad B_k &= \nabla_u f_\psi,
    &\quad r_k &= f_\psi - A_k \bar z_k^j - B_k \bar u_k^j, \\
C_k &= \nabla_z q_{\mathrm{safe}}, & D_k &= \nabla_u q_{\mathrm{safe}},
    & r_k' &= q_{\mathrm{safe}} - C_k \bar z_k^j - D_k \bar u_k^j, \\
g_k^z &= \nabla_z r_\theta, & g_k^u &= \nabla_u r_\theta,
    & g_N &= \nabla_z V_\phi(\bar z_N^j).
\end{aligned}
\end{equation}
We write $(z^{j+1}, u^{j+1})$ for the solution of~\eqref{eq:scp}. If it is accepted, the next reference is $\bar{u}^{j+1} = u^{j+1}$ together with its rollout $\bar{z}^{j+1}$ through $f_{\psi}$, so the reference always satisfies the dynamics exactly.
The seven Jacobians are computed
in parallel across the horizon, so an iteration's cost is dominated by the QP solve.

\paragraph{Trust region sizing.}\label{sec:tr_sizing} The proximal weight $w_{\mathrm{prox}}$ sets the size of the trust region: the larger its value, the closer the step must stay to its reference. After each sub-problem we compare the decrease in merit, namely the $\ell_1$-penalised objective of~\eqref{eq:ocp} evaluated on the rollout of the new plan, with the decrease the linearisation predicted \citep{mao2016successive},
\begin{equation}
    \rho_w = \frac{J_{\mathrm{nonlinear, old}} - J_{\mathrm{nonlinear, new}}}{J_{\mathrm{linear, old}} - J_{\mathrm{linear, new}}}.
\end{equation}
If $\rho_w \geq \rho_{\mathrm{lower}}$, the step is accepted, and if also $\rho_w > \rho_{\mathrm{upper}}$ the trust region weighting is loosened, $w_{\mathrm{prox}} \leftarrow \max(w_{\min}, w_{\mathrm{prox}}/2)$. If $\rho_w < \rho_{\mathrm{lower}}$, or the solver fails, the step is rejected and the trust region is tightened, $w_{\mathrm{prox}} \leftarrow 2 \cdot w_{\mathrm{prox}}$. 
We use $\rho_{\mathrm{lower}} = 0.25$, $\rho_{\mathrm{upper}} = 0.75$, an initial weight $w_{\mathrm{prox},0} = 1$ and
$w_{\min} = 10^{-2}$, and each planning step starts from the previous step's final weight, clipped to $[w_{\min}, 10]$.

\paragraph{Convergence criteria.} Let $\Delta^j = (z^{j+1}, u^{j+1}) - (\bar{z}^{j}, \bar{u}^{j})$ be the step taken by the sub-problem. Writing the proximal term as $\tfrac{1}{2t}\|\cdot\|_2^2$ with $t = 1/(2w_{\mathrm{prox}})$, the prox-gradient norm is $\|\Delta^j\|_2 / t = 2 w_{\mathrm{prox}} \|\Delta^j\|_2$. We stop when $2 \max(w_{\mathrm{prox}}, w_{\mathrm{prox},0}) \|\Delta^j\|_2 \le 10^{-2}$; taking the maximum with the initial weight stops a loosened trust region from shrinking the measure on its own. If the current plan's rollout then meets the barrier to
within $10^{-4}$ at every step, the plan is \emph{certified}; otherwise SCP has converged with barrier slack in use.
If neither happens within 40 iterations, SCP stops and returns the last accepted plan. Problem~\eqref{eq:ocp} with its constraints replaced by $\ell_1$ penalties is a convex function composed with smooth maps $f_{\psi}, r_{\theta}, V_{\phi}, q_{\mathrm{safe}}$, and~\eqref{eq:scp} is one step of the prox-linear method applied to it. When these maps have Lipschitz-continuous Jacobians (Proposition~\ref{prop:lin}) and $w_{\mathrm{prox}}$ is large enough, which the trust-region rule ensures, each step decreases the penalised objective and the prox-gradient norm tends to zero~\citep{lewis2016proximal, drusvyatskiy2019efficiency}. Our re-rollout of accepted plans is outside that analysis, but the acceptance test enforces the decrease directly. A point at which $\Delta^j = 0$ and the slack variables are zero is a KKT point of~\eqref{eq:ocp}, with multipliers bounded by the penalty weights \citep{mao2016successive}.

\paragraph{Initial guess.} As a local method, SCP converges to a point that depends on its start, which need not be safe. Plain SCP starts from its previous plan shifted by one step, its last action repeated (zeros at an episode's first
step). SCP+MPPI also runs a small MPPI planner (6 rounds of 512 sequences) \citep{bharadhwaj2020model} and starts from whichever of MPPI's plan and the shifted plan has the lower merit. Since only merit-decreasing steps are accepted, its plan is never worse under the merit than MPPI's, whether or not it converges.

\section{Experimental Setup}
We ask: (Q1) whether at a matched wall-clock budget, SCP from a sampled initialisation finds safer and higher-return plans than sampling-based planning; (Q2) how this changes as the feasible set narrows through a larger safety margin, a longer horizon, or more hazards; (Q3) how often SCP converges within its iteration cap, how this depends on the regularity constraint, and what a converged plan tells us that a sampled one does not.

\paragraph{Environments.} We use SafetyPointGoal1 and SafetyPointGoal2 from Safety-Gymnasium \citep{ji2023safety}. A point robot with two actuators (forward force and turning, $u \in [-1, 1]^2)$, must reach a goal within $T=1000$ steps. The goal moves to a new position each time it is reached. The reward is the per-step reduction in distance to the goal, plus a bonus of $1$ for reaching it.
The agent observes 12 proprioceptive signals and three 16-bin lidars measuring the proximity of the goal, hazards and vases. PointGoal1 has 8 hazards and one cost-free vase. PointGoal2 has 10 hazards and 10 vases, and contact with vases is also now a violation. A step is a violation if it incurs any cost. We treat violations as absorbing failures, hence an episode will end at its first violation. The violation rate is therefore the share of episodes that fail, and the return is the reward accumulated before the end of the episode. The action is two-dimensional, which favours sampling-based methods as the search space is small, making this a demanding comparison for successive convexification.

\paragraph{Offline data and world models.} For each task we train on the corresponding DSRL
dataset~\citep{liu2023dsrl}: 2,022 trajectories for PointGoal1 and 3,442 for PointGoal2, each of 1000 steps. We
truncate every trajectory at its first violation, which becomes a terminal transition, to match the absorbing failure
above. This leaves about 0.88\,M transitions for PointGoal1 and 1.15\,M for PointGoal2. Only 329 of PointGoal2's trajectories run their full length without a violation. We train one model per task and bound, $c \in \{1.2, 1.5, \infty\}$ ($\infty$: weights not projected), at $N = 10$, and two more PointGoal1 models with $c = 1.5$ for $N = 20$ and 30 (details in Appendix~B).

\paragraph{Planners.} All planners solve~\eqref{eq:ocp} in the same model and replan at every step. SCP and SCP+MPPI start as in Section~\ref{sec:controlstuff} and stop after at most 40 iterations, to prevent crawling (many iterations of vanishing progress)~\citep{uzun2024successive}. The baselines are MPPI~\citep{williams2018information}
and CEM~\citep{rubinstein1999cross}, which run 6 rounds of $J$ sequences drawn from a Gaussian, refitted each round to
the best $J/8$ (weighted by $\exp(\text{score}/0.5)$ in MPPI, uniformly in CEM) and warm-started from the previous mean
shifted by one step as in TD-MPC \citep{hansen2022temporal}. The samplers enforce the same barrier through ranking violating trajectories below those that remain safe \citep{wen2018constrained}, and if none remain safe, sequences are ranked by their total barrier excess, i.e. the quantity
SCP's $\ell_1$ penalty charges. They execute the refitted mean if its rollout meets the barrier, and otherwise the best sampled sequence. We match the samplers' time per step to SCP+MPPI's (Appendix~B): $J = 65{,}536$, 95,040 and 120,000 sequences per round at $N = 10$, 20 and 30.

\paragraph{Evaluation.} We sweep the margin $\epsilon \in \{ 0.0, 0.1, 0.2, 0.3, 0.5 \}$m on every $N=10$ model, and fix $\epsilon=0.2$ for the horizon study. Each configuration is run for 30 episodes, with seeds shared between planners so that every planner faces the same 30 layouts. We report violation rates with Wilson $95 \%$ intervals, and mean returns with bootstrap $95 \%$ intervals ($10^4$ resamples). To compare planners we pair episodes on world model, margin and seed. The main comparison pools the two $c = 1.5$ models and the four margins $\epsilon \ge 0.1$, which gives 240 pairs (at $\epsilon = 0$ every planner fails often); Appendix~C repeats it for $c = 1.2$ and $c = \infty$. For each comparison we report the difference in violation rate and in return with 95\% intervals, and the probability, given the data, that SCP+MPPI's violation rate is lower, $P(\text{safer})$, or its return higher, $P(\text{higher return})$, in the spirit of \citet{agarwal2021deep}. Appendix~C gives the methods, chosen to stay sound when violations are rare.

\section{Results}

\subsection{Safety and Planning Quality at Matched Wall-Clock Time}

\begin{table}[t]
  \centering\footnotesize
  \begin{tabular}{@{}llrcccc@{}}
\toprule
 & Baseline & Pairs & $\Delta$ violations (pp) & $P$(safer) & $\Delta$ return & $P$(higher return) \\
\midrule
Main & SCP & 240 & $-36.7$ [$-43.8$, $-28.8$] & $>0.999$ & $+0.17$ [$-0.05$, $+0.40$] & 0.94 \\
 & MPPI & 240 & $-3.8$ [$-9.0$, $+1.5$] & 0.92 & $+0.07$ [$-0.13$, $+0.25$] & 0.78 \\
 & CEM & 240 & $-3.3$ [$-8.4$, $+1.7$] & 0.90 & $+0.18$ [$+0.06$, $+0.31$] & 0.999 \\
\midrule
$N = 10$ & SCP & 30 & $-40.0$ [$-57.9$, $-16.1$] & 0.999 & $+0.49$ [$-0.05$, $+1.02$] & 0.96 \\
 & MPPI & 30 & $-16.7$ [$-29.7$, $-3.2$] & 0.98 & $-0.10$ [$-0.61$, $+0.33$] & 0.36 \\
 & CEM & 30 & $-10.0$ [$-21.4$, $+1.0$] & 0.94 & $+0.21$ [$-0.17$, $+0.57$] & 0.88 \\
\midrule
$N = 20$ & SCP & 30 & $-26.7$ [$-44.7$, $-7.3$] & 0.994 & $+1.49$ [$+0.70$, $+2.20$] & $>0.999$ \\
 & MPPI & 30 & $-3.3$ [$-15.4$, $+6.9$] & 0.75 & $+0.65$ [$+0.02$, $+1.29$] & 0.98 \\
 & CEM & 30 & $-3.3$ [$-15.4$, $+6.9$] & 0.75 & $+0.93$ [$+0.15$, $+1.71$] & 0.99 \\
\midrule
$N = 30$ & SCP & 30 & $-20.0$ [$-36.9$, $-5.6$] & 0.992 & $+1.15$ [$+0.01$, $+2.30$] & 0.98 \\
 & MPPI & 30 & $+3.3$ [$-8.3$, $+16.7$] & 0.25 & $+1.13$ [$+0.40$, $+1.90$] & 0.998 \\
 & CEM & 30 & $+3.3$ [$-8.3$, $+16.7$] & 0.25 & $+0.98$ [$+0.07$, $+1.91$] & 0.98 \\
\bottomrule
\end{tabular}
  \caption{SCP+MPPI minus SCP and each time-matched sampler. Main: the main comparison ($c = 1.5$, $N = 10$, both tasks, $\epsilon \ge 0.1$). $N = 10$, 20, 30: PointGoal1 at each horizon ($c = 1.5$, $\epsilon = 0.2$; 30 pairs), where at $N = 20$ and 30 there are too few violations to compare safety. Negative $\Delta$ violations and positive $\Delta$ return favour SCP+MPPI; brackets are 95\% intervals.}
  \label{tab:baseline_comparison_N10}\label{tab:horizon_comparison}
\end{table}
In Table~\ref{tab:baseline_comparison_N10} the estimates favour SCP+MPPI on both measures, but only one of the four differences is resolved. Its violation rate is $21 \%$ against $25 \%$ for each sampler, $3.8$ and $3.3$ points lower than MPPI and CEM respectively, with intervals that include zero. The probability that SCP+MPPI is the safer planner is 0.92 against MPPI, and 0.90 against CEM. Its return is higher than MPPI's by $0.07$ (not resolved) and than CEM's by 0.18, where the interval excludes zero.
This lean is specific to $c = 1.5$: over all six $N = 10$ models the differences from time-matched MPPI and CEM are $-0.1$ and $+0.7$ points; neither is resolved (Appendix~C).
The sampled start matters: SCP started from its previous plan fails in $58 \%$ of episodes against $21 \%$ (Table~\ref{tab:baseline_comparison_N10}), and is less safe on each of the six models.\looseness=-1

\begin{figure}[t]
    \centering
    \includegraphics[width=\linewidth]{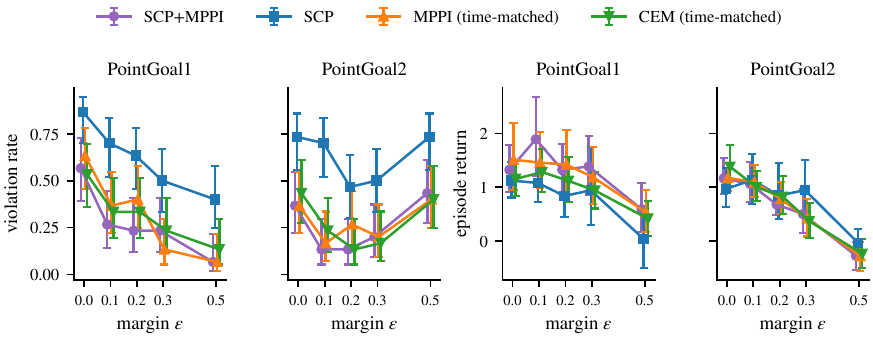}
    \caption{Violation rate (left pair) and return (right pair) against the margin $\epsilon$ on each task’s $c = 1.5$ model at $N = 10$, with $95 \%$ intervals.}
    \label{fig:eps_sweep_1_5}
\end{figure}

\subsection{Narrowing the Feasible Set}
\paragraph{Safety margin and hazards.} On PointGoal1 (Fig.~\ref{fig:eps_sweep_1_5}) a larger margin makes every planner safer. The violation rate of SCP+MPPI falls from $57 \%$ at $\epsilon = 0$ to $7 \%$ at $\epsilon = 0.5$, and its return peaks at 1.9 at $\epsilon = 0.1$ and falls to 0.6 at $\epsilon=0.5$. On PointGoal2, which adds hazards and vases, the margin helps only up to a point. The violation rate falls from $37 \%$ to $13 \%$ at $\epsilon = 0.1$ and $0.2$, then rises to $43 \%$ at $\epsilon = 0.5$, where the mean return is negative. On five of the six models, the violation rate of SCP+MPPI is lowest at $\epsilon = 0.1$ or $0.2$ (Fig.~\ref{fig:eps_sweep_viol}). The time-matched samplers follow the same curves. Of their 20 per-margin differences from SCP+MPPI, only one is resolved, in its favour (PointGoal1, $\epsilon = 0.2$, MPPI, Table~\ref{tab:horizon_comparison}).

\paragraph{Horizon.} A longer horizon makes every planner safer, and it is where the returns separate (Table \ref{tab:horizon_comparison}). Violations are too few at $N = 20$ and 30 to compare safety, but SCP+MPPI earns $0.65$ and $0.93$ more than time-matched MPPI and CEM at $N=20$, and $1.13$ and $0.98$ more at $N=30$. All four intervals exclude zero. Each $N$ uses its own world model, trained on windows of $N + 1$ steps, so the effect of the horizon cannot be separated from that of the world model.

\subsection{Convergence to a Karush-Kuhn-Tucker Point, and What It Buys}

\begin{figure}[t]
    \centering
    \includegraphics[width=\linewidth]{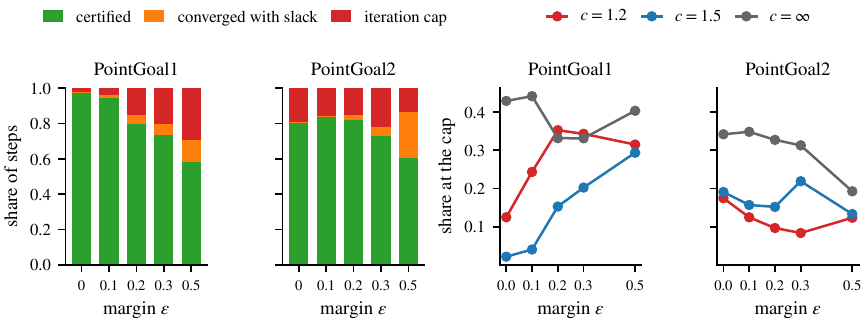}
    \caption{SCP+MPPI's convergence at $N = 10$. Left pair: how its steps ended on each task's $c = 1.5$ model (certified: converged with the plan's rollout meeting the barrier). Right pair: the share of its steps at the 40-iteration cap for each Lipschitz bound.}
    \label{fig:convergence}
\end{figure}

A \emph{certified} plan (Section~\ref{sec:controlstuff}) has zero slack, because the dynamics hold exactly (the reference is always the rollout of its plan), so it is an approximate KKT point of the OCP, feasible in the model, and no first-order change improves it. On the $c=1.5$ models at $N=10$, SCP+MPPI certifies $97 \%$ of its steps at $\epsilon = 0$ and $58 \%$ at $\epsilon = 0.5$ on PointGoal1, and $81 \%$ and $61 \%$ on PointGoal2.
\
\paragraph{The learned barrier is a reliable monitor.} At $\epsilon \ge 0.1$, a violation follows within $N$ steps of a plan that meets the barrier in the model on $4 \times 10^{-4}$ of SCP+MPPI's steps, against $3 \times 10^{-2}$ after one that does not, and 93--98\% of each planner's violations follow a plan that failed the barrier (Fig.~\ref{fig:crash_types}). MPPI and CEM show the same gap, so this is a property of the learned barrier, not of SCP. The model is optimistic about return for every planner, but no more so under SCP+MPPI (Appendix~D).

\paragraph{What SCP adds.} Beyond the sampled plan, SCP gives first-order optimality in the model when it converges (on 53\% and 32\% of steps at $N = 20$ and 30), and infeasibility detection. Converging with slack still in use indicates that no feasible plan lies near the current one, provided the penalty weights exceed the multipliers. That share grows as the safe set shrinks, from 1\% to 12\% of steps on PointGoal1 and 0\% to 26\% on PointGoal2, where at $\epsilon = 0.5$ it coincides with the rise in violations. A sampler finding no feasible sequence cannot tell an empty safe set from too few samples.

\paragraph{Effect of the Lipschitz bound.} Bounding the Lipschitz constant lets SCP converge within 40 iterations more often (Fig.~\ref{fig:convergence}). On the unbounded models SCP+MPPI stops at the cap on 33--44\% of steps on PointGoal1 and 19--35\% on PointGoal2; with $c=1.5$, on 2--29\% and 13--22\%. A bounded model reaches the cap less often than the unbounded one in $18$ of the $20$ comparisons across tasks and margins, and the bound did not cost return (Fig. \ref{fig:eps_sweep_returns}). With one trained model per task and bound, however, we cannot separate the effect of c from variation between runs.

\section{Limitations and Conclusion}
\label{sec:conclusion}

We posed safe planning in a learned latent world model as a successive convex program with a barrier constraint, with regularity conditions that bound its linearisation error. From a small sampled plan, SCP violated less often than MPPI and CEM at equal wall-clock time in the main comparison, earned greater return, and labelled each step certified, infeasible or not converged.

Several limitations remain. Safety is enforced in the learned model and not in
the environment. Between a fifth and a quarter of episodes still end in a
violation, most of them after a plan that the model itself flagged as unsafe,
and we have no fallback for such steps. Planning takes 0.4 to 0.7\,s per step at $N = 10$ and 2.2 to 3.6\,s at the longer horizons, and at $N = 20$ samplers with 512 sequences reach a similar return far faster. The evidence covers two tasks with
two-dimensional actions, one trained model per setting and 30 episodes per configuration, and the horizon study changes the model with the horizon. Finally, the convergence guarantee does not cover our rollout of each accepted plan; reading convergence with slack as
local infeasibility assumes penalty weights above the constraint multipliers; and the barrier needs distance measurements in the observation. The natural next steps are using the infeasibility signal to trigger a fallback controller, and testing on
higher-dimensional action spaces.

\subsection*{Generative AI Use}
Generative AI tools were used for literature search and experimental implementation. The author reviewed and verified all AI-assisted content and takes responsibility for the final content of the paper.

\acks{The author thanks Lukas Talmacsi-Marby for helpful contributions.}

\bibliography{l4dc2027-sample}

\newpage
\appendix

\section{Supplementary Mathematics}\label{app:math}
\subsection{Lipschitz Constant of an Affine Layer}\label{app:affine-lip}
Let $T(\vec{x}) = W\vec{x} + \vec{b}$. Its Lipschitz constant with respect
to $\|\cdot\|_p$ is
\[
  L(T) = \sup_{\vec{x}_1 \neq \vec{x}_2}
  \frac{\| T(\vec{x}_1) - T(\vec{x}_2) \|_p}{\| \vec{x}_1 - \vec{x}_2 \|_p} .
\]
The bias cancels, since
$T(\vec{x}_1) - T(\vec{x}_2) = W(\vec{x}_1 - \vec{x}_2)$.
Setting $\vec{a} = \vec{x}_1 - \vec{x}_2$, which ranges over all nonzero
vectors as $\vec{x}_1 \neq \vec{x}_2$ vary,
\[
  L(T) = \sup_{\vec{a} \neq \vec{0}} \frac{\| W\vec{a} \|_p}{\| \vec{a} \|_p}
  = \|W\|_p ,
\]
yielding that the bound on the Lipschitz norm of any fully connected neural network layer is given by the $p$-norm of $W$.

\subsection{Bounding the Lipschitz Constant of a Network}\label{app:network-lip}
Consider a network
$f = T_d \circ \sigma \circ T_{d-1} \circ \cdots \circ \sigma \circ T_1$
with affine layers $T_i(\vec{x}) = W_i\vec{x} + \vec{b}_i$ and activations
$\sigma_i$. Since
$L(g \circ h) \leq L(g)\,L(h)$ for any Lipschitz $g, h$,
\[
  L(f) \leq \prod_{i=1}^{d} L(T_i) \prod_{i=1}^{d-1} L(\sigma_i) = \prod_{i=1}^{d} \|W_i\|_p \prod_{i=1}^{d-1} L(\sigma_i).
\]
We measure distances in the $\ell_2$ norm, so $\|W_i\|_2$ is the spectral norm
(largest singular value) of $W_i$. To enforce $\|W_i\|_2 \leq c_i$, we rescale
the weights after each update by a projection function
\[
  \mathrm{proj}(W,c) = \frac{W}{\max\bigl(1, \frac{\|W\|_2}{c}\bigr)} ,
\]
which leaves $W$ unchanged if $\|W\|_2 \leq c$ and otherwise scales it to
$\|\mathrm{proj}(W,c)\|_2 = c$. Hence $L(f) \leq \prod_{i=1}^{d} c_i L(\sigma_i)$.

\subsection{Proof of Proposition~\ref{prop:lin}}
\label{app:prop1}
 
Throughout, $\|\cdot\|_2$ is the Euclidean norm of a vector and the spectral norm of a matrix, $\nabla$ denotes a Jacobian, and $L(g)$ is the Lipschitz constant of $g$. We use two standard facts. First, if $g$ is continuously differentiable and $\|\nabla g(x)\|_2 \le M$ for all $x$, then $g$ is $M$-Lipschitz. Second, $L(g \circ h) \le L(g)\, L(h)$.
 
\begin{proof}
\textbf{Part (i).} The activation $\sigma_\delta$ acts on each pair $(a_i, b_i)$ separately, mapping it to
\[
  u_i = \tfrac12\big(a_i + b_i + \phi_\delta(a_i - b_i)\big),
  \qquad
  v_i = \tfrac12\big(a_i + b_i - \phi_\delta(a_i - b_i)\big).
\]
Write $t = a_i - b_i$. Since $\phi_\delta(t) = \sqrt{t^2 + \delta^2}$, we have $\phi_\delta'(t) = t / \sqrt{t^2 + \delta^2}$, and so $|\phi_\delta'(t)| < 1$. Differentiating $u_i$ and $v_i$ with respect to $a_i$ and $b_i$, the Jacobian of the map $(a_i, b_i) \mapsto (u_i, v_i)$ is
\[
  G(t) = \frac12
  \begin{pmatrix}
    1 + \phi_\delta'(t) & 1 - \phi_\delta'(t) \\
    1 - \phi_\delta'(t) & 1 + \phi_\delta'(t)
  \end{pmatrix}
  = P_+ + \phi_\delta'(t)\, P_- ,
\]
where
\[
  P_+ = \frac12 \begin{pmatrix} 1 & 1 \\ 1 & 1 \end{pmatrix},
  \qquad
  P_- = \frac12 \begin{pmatrix} 1 & -1 \\ -1 & 1 \end{pmatrix}.
\]
The matrices $P_+$ and $P_-$ are the orthogonal projectors onto the directions $(1, 1)$ and $(1, -1)$. Hence $G(t)$ is symmetric with eigenvalues $1$ and $\phi_\delta'(t)$, and $\|G(t)\|_2 = \max\{1, |\phi_\delta'(t)|\} = 1$.
 
After reordering the coordinates so that the two members of each pair are adjacent, the Jacobian of $\sigma_\delta$ at $x$ is block diagonal with blocks $G(t_1), \dots, G(t_k)$, where $t_i = a_i - b_i$. Reordering coordinates does not change the spectral norm, and the spectral norm of a block-diagonal matrix is the largest norm among its blocks. Therefore $\|\nabla \sigma_{\delta}(x)\|_2 = 1$ for every $x$, and $\sigma_\delta$ is 1-Lipschitz by the first fact above.
 
Each affine layer has $L(T_i) = \|W_i\|_2 \le c_i$ (Appendix~\ref{app:affine-lip}). Applying the second fact to $f = T_n \circ \sigma_\delta \circ \dots \circ \sigma_\delta \circ T_1$ gives $L(f) \le \prod_{i=1}^{n} c_i$.
 
\medskip
\textbf{Part (ii).} Differentiating $\phi_\delta'$ once more,
\[
  \phi_\delta''(t) = \frac{1}{\sqrt{t^2 + \delta^2}} - \frac{t^2}{(t^2 + \delta^2)^{3/2}} = \frac{\delta^2}{(t^2 + \delta^2)^{3/2}} ,
\]
which is positive and largest at $t = 0$, where it equals $1/\delta$. So $\phi_\delta'$ is $(1/\delta)$-Lipschitz:
\[
  |\phi_\delta'(s) - \phi_\delta'(t)| \le \frac{1}{\delta}\, |s - t| \qquad \text{for all } s, t .
\]
Now take two inputs $x$ and $y$, and write $t_i^x$ and $t_i^y$ for the differences $a_i - b_i$ of their $i$-th pairs. In the reordered coordinates of Part (i), the term $P_+$ cancels in the difference of the two Jacobians, which is therefore block diagonal with blocks $\big(\phi_\delta'(t_i^x) - \phi_\delta'(t_i^y)\big) P_-$. Since $\|P_-\|_2 = 1$,
\[
  \|\nabla\sigma_{\delta}(x) - \nabla\sigma_{\delta}(y)\|_2
  = \max_i \big| \phi_\delta'(t_i^x) - \phi_\delta'(t_i^y) \big|
  \le \frac{1}{\delta} \max_i |t_i^x - t_i^y| .
\]
Finally, $t_i^x - t_i^y = (a_i^x - a_i^y) - (b_i^x - b_i^y)$ is the inner product of $(1, -1)$ with the change in the $i$-th pair, so by the Cauchy--Schwarz inequality
\[
  |t_i^x - t_i^y| \le \sqrt{2}\, \big\| (a_i^x - a_i^y,\ b_i^x - b_i^y) \big\|_2 \le \sqrt{2}\, \|x - y\|_2 .
\]
Combining the last two displays, $\|\nabla\sigma_{\delta}(x) - \nabla\sigma_{\delta}(y)\|_2 \le (\sqrt{2}/\delta)\, \|x - y\|_2$.
 
\medskip
\textbf{Part (iii).} For $i = 1, \dots, n - 1$, let
\[
  f_i = T_i \circ \sigma_\delta \circ \dots \circ \sigma_\delta \circ T_1
\]
be the network up to its $i$-th affine layer, so that $f = T_n \circ \sigma_\delta \circ f_{n-1}$. As $f_i$ is built from the first $i$ affine layers and $i - 1$ activations, Part (i) gives $L(f_i) \le \prod_{j \le i} c_j$. Let $S_i(x) = \nabla\sigma_\delta(f_i(x))$ be the Jacobian of the $i$-th activation, evaluated at the input that $x$ produces for it. By the chain rule,
\[
  \nabla f(x) = W_n\, S_{n-1}(x)\, W_{n-1}\, S_{n-2}(x) \cdots S_1(x)\, W_1 .
\]
The weight matrices do not depend on the input, so the Jacobians at $x$ and at $y$ differ only in the factors $S_1, \dots, S_{n-1}$. Changing these factors from $y$ to $x$ one at a time gives the telescoping sum
\[
  \nabla f(x) - \nabla f(y) = \sum_{i=1}^{n-1} W_n\, S_{n-1}(x) \cdots W_{i+1} \big[ S_i(x) - S_i(y) \big] W_i\, S_{i-1}(y) \cdots W_1 ,
\]
in which the factors to the left of the bracket are evaluated at $x$ and those to its right at $y$. We bound the norm of the $i$-th term by the product of the norms of its factors. Every weight matrix appears once, contributing at most $\prod_{j=1}^{n} c_j$ in total. Every $S_j$ outside the bracket has norm $1$ by Part (i). For the bracket, Part (ii) and then the bound on $L(f_i)$ give
\[
  \|S_i(x) - S_i(y)\|_2
  \le \frac{\sqrt{2}}{\delta}\, \|f_i(x) - f_i(y)\|_2
  \le \frac{\sqrt{2}}{\delta} \Big( \prod_{j \le i} c_j \Big) \|x - y\|_2 .
\]
The norm of the $i$-th term is therefore at most
\[
  \Big( \prod_{j=1}^{n} c_j \Big)\, \frac{\sqrt{2}}{\delta}\, \Big( \prod_{j \le i} c_j \Big)\, \|x - y\|_2 ,
\]
and summing over $i = 1, \dots, n - 1$,
\[
  \|\nabla f(x) - \nabla f(y)\|_2 \le \frac{\sqrt{2}}{\delta} \Big( \prod_{j=1}^{n} c_j \Big) \sum_{i=1}^{n-1} \prod_{j \le i} c_j \ \|x - y\|_2 ,
\]
which is the stated bound on $\beta$.
 
\medskip
\textbf{The linearisation bound.} Since $f$ is continuously differentiable, the fundamental theorem of calculus along the segment from $x$ to $x + \Delta$ gives $f(x + \Delta) - f(x) = \int_0^1 \nabla f(x + \xi\Delta)\, \Delta \, \mathrm{d}\xi$. Subtracting $\nabla f(x)\Delta = \int_0^1 \nabla f(x)\, \Delta \, \mathrm{d}\xi$,
\[
  f(x + \Delta) - f(x) - \nabla f(x)\Delta = \int_0^1 \big( \nabla f(x + \xi\Delta) - \nabla f(x) \big) \Delta \, \mathrm{d}\xi .
\]
By Part (iii), $\|\nabla f(x + \xi\Delta) - \nabla f(x)\|_2 \le \beta\, \xi\, \|\Delta\|_2$, so the norm of the right-hand side is at most $\int_0^1 \beta\, \xi\, \|\Delta\|_2^2 \, \mathrm{d}\xi = \tfrac{\beta}{2} \|\Delta\|_2^2$.
\end{proof}
 
\paragraph{Remark (the bound is a worst case).}
The constant in Part (ii) is approached only where a pair of pre-activations is nearly tied, since $\phi_\delta''$ falls from $1/\delta$ at $t = 0$ to about a third of that at $|t| = \delta$. For three layers with $\prod_i c_i = 1.5$ split evenly and $\delta = 0.05$, the proposition gives $\beta \le 104$.
 
\paragraph{Remark (compositions).}
The barrier constraint composes two networks, $g_\Phi \circ f_\psi$. If $f$ is $L_f$-Lipschitz with a $\beta_f$-Lipschitz Jacobian, and $g$ is $L_g$-Lipschitz with a $\beta_g$-Lipschitz Jacobian, then the Jacobian of $g \circ f$ is $(\beta_g L_f^2 + L_g \beta_f)$-Lipschitz. This follows from the chain rule, $\nabla (g \circ f)(x) = \nabla g(f(x))\, \nabla f(x)$, and the splitting
\begin{align*}
  \nabla g(f(x))\, \nabla f(x) - \nabla g(f(y))\, \nabla f(y)
  ={} & \big[ \nabla g(f(x)) - \nabla g(f(y)) \big] \nabla f(x) \\
  & + \nabla g(f(y)) \big[ \nabla f(x) - \nabla f(y) \big] ,
\end{align*}
whose two terms have norms at most $\beta_g L_f \|x - y\|_2 \cdot L_f$ and $L_g \cdot \beta_f \|x - y\|_2$. The proposition therefore also bounds the linearisation error of the constraint, and the same argument applied along the horizon bounds that of the rollout.

\subsection{Derivation of the Convex Sub-problem}
\label{app:scp-derivation}
 
Problem~\eqref{eq:ocp} is non-convex in two places: its objective contains the
learned maps $r_\theta$ and $V_\phi$, and its constraints contain $f_\psi$ and $q_{\mathrm{safe}}$. We obtain
\eqref{eq:scp} from it in three steps.
 
\paragraph{Step 1: penalty.}
Constraints that are linearised directly can be inconsistent even when
\eqref{eq:ocp} is feasible \citep{mao2016successive}. We therefore move the two learned
constraints into the objective as $\ell_1$ penalties with weights
$w_{\mathrm{dyn}}, w_{\mathrm{safe}} > 0$, which gives the merit function
\begin{equation}
\begin{aligned}
J_{\mathrm{nonlinear}}(z,u) ={}&
  \sum_{k=0}^{N-1}\bigl(\rho_u\|u_k\|_2^2-\gamma^k r_\theta(z_k,u_k)\bigr)
  -\gamma^N V_\phi(z_N)\\
&+w_{\mathrm{dyn}}\sum_{k=0}^{N-1}\bigl\|z_{k+1}-f_\psi(z_k,u_k)\bigr\|_1\\
&+w_{\mathrm{safe}}\sum_{k=0}^{N-1}\max\bigl\{0,\,q_{\mathrm{safe}}(z_k,u_k)\bigr\},
\end{aligned}
\label{eq:merit}
\end{equation}
to be minimised over $z_{1:N}$ and $u_{0:N-1}$ subject to $u_k\in\mathcal{U}$
and $z_0=z_t$. This is the merit used in the ratio test (Section \ref{sec:tr_sizing}).

On a rollout the dynamics term is zero, so there the merit is the objective of
\eqref{eq:ocp} plus the barrier penalty.
 
\paragraph{Step 2: linearisation.}
Let $(\bar z^j,\bar u^j)$ be the reference. We replace $f_\psi$,
$q_{\mathrm{safe}}$, $r_\theta$ and $V_\phi$ in \eqref{eq:merit} by their
first-order expansions about it, using the terms of \eqref{eq:scp-lin}, and
discard the constants
$r_\theta(\bar z_k^j,\bar u_k^j)-(g_k^z)^\top\bar z_k^j-(g_k^u)^\top\bar u_k^j$
and $V_\phi(\bar z_N^j)-g_N^\top\bar z_N^j$, which do not change the
minimiser. This gives
\begin{equation}
\begin{aligned}
J_{\mathrm{linear}}(z,u) ={}&
  \sum_{k=0}^{N-1}\Bigl(\rho_u\|u_k\|_2^2
  -\gamma^k\bigl((g_k^z)^\top z_k+(g_k^u)^\top u_k\bigr)\Bigr)
  -\gamma^N g_N^\top z_N\\
&+w_{\mathrm{dyn}}\sum_{k=0}^{N-1}\bigl\|z_{k+1}-A_kz_k-B_ku_k-r_k\bigr\|_1\\
&+w_{\mathrm{safe}}\sum_{k=0}^{N-1}\max\bigl\{0,\,C_kz_k+D_ku_k+r_k'\bigr\}.
\end{aligned}
\label{eq:merit-lin}
\end{equation}

The sub-problem adds a proximal term, which keeps the step where the expansion
is accurate:
\begin{equation}
\min_{z_{1:N},\,u_{0:N-1}}\;
J_{\mathrm{linear}}(z,u)
+w_{\mathrm{prox}}\Bigl(\sum_{k=1}^{N}\|z_k-\bar z_k^j\|_2^2
+\sum_{k=0}^{N-1}\|u_k-\bar u_k^j\|_2^2\Bigr)
\quad\text{s.t.}\quad u_k\in\mathcal{U},\ \ z_0=z_t .
\label{eq:proxlin}
\end{equation}
Every term of \eqref{eq:proxlin} is convex: a convex quadratic, a linear
function, and an $\ell_1$ norm and a hinge of affine functions. The proximal
term makes the objective strongly convex in $(z_{1:N},u)$, so its minimiser is
unique. Problem~\eqref{eq:proxlin} is one step of the prox-linear method
\citep{lewis2016proximal,drusvyatskiy2019efficiency} applied to \eqref{eq:merit}, with step
$t=1/(2w_{\mathrm{prox}})$.
 
\paragraph{Step 3: quadratic program.}
Problem~\eqref{eq:proxlin} is convex but not smooth. Its two non-smooth terms
have linear-programming forms: for $d\in\mathbb{R}^{n_z}$ and
$s\in\mathbb{R}$,
\begin{align}
\|d\|_1 &= \min_{\mu^+,\,\mu^-}\bigl\{\mathbf{1}^\top(\mu^++\mu^-) :
  \mu^+-\mu^-=d,\ \mu^+\ge 0,\ \mu^-\ge 0\bigr\},
  \label{eq:l1-lp}\\
\max\{0,s\} &= \min_{\nu}\bigl\{\nu : \nu\ge s,\ \nu\ge 0\bigr\},
  \label{eq:hinge-lp}
\end{align}
with the minima attained at $\mu^+=\max\{d,0\}$, $\mu^-=\max\{-d,0\}$
(element-wise) and $\nu=\max\{0,s\}$. We substitute \eqref{eq:l1-lp} with
$d=z_{k+1}-A_kz_k-B_ku_k-r_k$ and \eqref{eq:hinge-lp} with
$s=C_kz_k+D_ku_k+r_k'$ into \eqref{eq:proxlin}. Because
$w_{\mathrm{dyn}},w_{\mathrm{safe}}>0$ and each new variable appears in a
single term, the inner minimisations can be merged with the outer one, and the
result is exactly \eqref{eq:scp}. Its objective is quadratic and, since
$\mathcal{U}$ is a box, its constraints are linear, so \eqref{eq:scp} is a QP.

\section{Experimental Details}\label{app:details}

\subsection{Environments}\label{app:env}
\begin{figure}[htbp]
    \centering
    \includegraphics[width=0.62\linewidth]{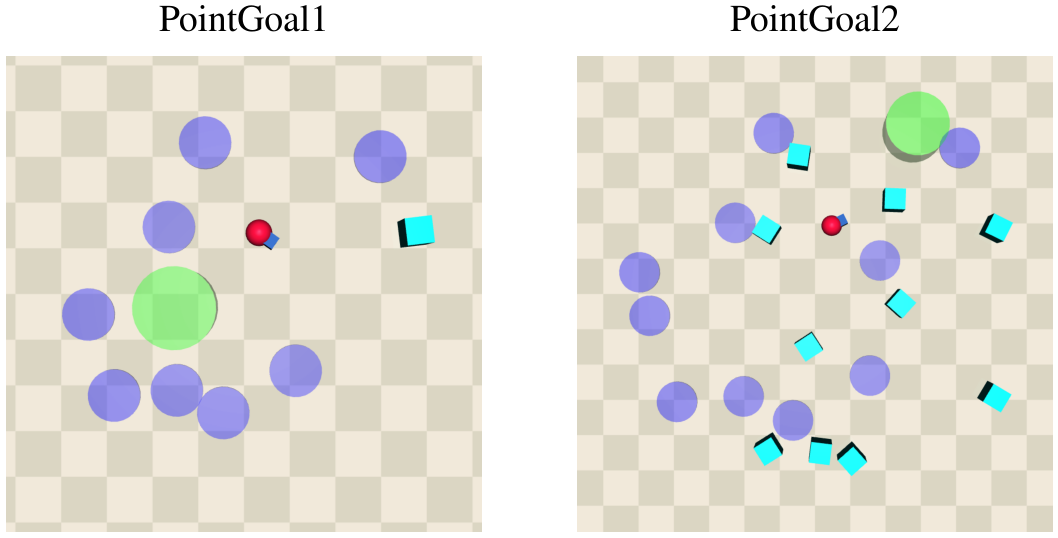}
    \caption{The two tasks from above, one layout each (layouts are drawn from the episode's seed). The point robot (red) must reach the goal (green) without entering a hazard (blue). In PointGoal2, touching a vase (cyan) is also a violation.}
    \label{fig:envs}
\end{figure}

SafetyPointGoal1 and SafetyPointGoal2~\citep{ji2023safety} are shown in Fig.~\ref{fig:envs}. Level~1 places 8 hazards of radius 0.2\,m and one cost-free vase in a $3 \times 3$\,m region, level~2 places 10 hazards and 10 vases in a $4 \times 4$\,m region. The goal has radius 0.3\,m and is moved after each visit, episodes last at most 1000 steps and end at the first violation. The layout, the initial state and the planners' random numbers are fixed by the seed.

\subsection{World models}
Table~\ref{tab:wm} lists the world-model settings.

\begin{table}[htbp]
\centering\small
\begin{tabular}{@{}p{0.27\linewidth}p{0.69\linewidth}@{}}
\toprule
Latent dimension $n_z$ & 32 \\
Encoder input & 53 of the 60 observation channels (drops the accelerometer and the out-of-plane velocity,
angular-rate and magnetometer components) \\
Dynamics, reward, value, barrier & MLP, $2 \times 256$, smoothed MaxMin ($\delta = 0.05$), the network bound $c$ is
split evenly over the three layers ($c^{1/3}$ each), the dynamics network is residual, $z + g_\psi(z, u)$ with $g_\psi$
bounded by $c - 1$ \\
$Q_\Theta$ (training only) & MLP, $2 \times 256$, unbounded \\
Reward target & dense part of the reward, $r - \mathds{1}[r > 0.5]$, the goal bonus is left to the value \\
Optimiser & Adam, learning rate $3 \times 10^{-4}$, gradient-norm clip 20 \\
Training & 250 epochs of 1000 (PointGoal1) or 1800 (PointGoal2) minibatches of 256 windows \\
Window, step weight & $N + 1$ transitions, $\rho = 0.9, 0.95, 0.99$ for $N = 10, 20, 30$ \\
Loss weights & dynamics 1, reward 1, value 0.5, barrier 1, VICReg $\gamma_v = 0.5$, $\lambda_{\mathrm{var}} = 1$,
$\lambda_{\mathrm{cov}} = 0.04$ \\
Discount, expectile & $\gamma = 0.99$, $\tau = 0.7$ \\
Target networks & moving averages with rates 0.01 (encoder) and 0.005 ($Q_\Theta$) \\
Barrier targets & lidar distance to the nearest hazard edge (PointGoal2: hazard or vase), contact distance 0.2\,m
(hazards) and 0.3\,m (vases), clipped at 1\,m; rows weighted $1 + 4\exp(-\mathrm{dist}/0.2)$ \\
Validation & 10\% of trajectories held out \\
\bottomrule
\end{tabular}
\caption{World-model settings. One model is trained per task, bound and horizon, with seed 0.}
\label{tab:wm}
\end{table}

\subsection{Planners}
Table~\ref{tab:planners} lists the planner settings.

\begin{table}[htbp]
\centering\small
\begin{tabular}{@{}p{0.2\linewidth}p{0.38\linewidth}p{0.36\linewidth}@{}}
\toprule
 & SCP and SCP+MPPI & MPPI and CEM \\
\midrule
Shared & \multicolumn{2}{p{0.74\linewidth}}{$N \in \{10, 20, 30\}$, $\rho_u = 10^{-4}$, $\alpha = 0.05$,
$\epsilon \in [0, 0.5]$\,m} \\
Iterations & at most 40 & 6 rounds \\
Penalties & $w_{\mathrm{dyn}} = 10^3$, $w_{\mathrm{safe}} = 10^2$ & -- \\
Trust region & $w_{\mathrm{prox},0} = 1$, $w_{\min} = 10^{-2}$, $(\rho_{\mathrm{lower}}, \rho_{\mathrm{upper}}) =
(0.25, 0.75)$ & -- \\
Stopping & prox-gradient $10^{-2}$, feasibility $10^{-4}$ & -- \\
QP solver & Clarabel & -- \\
Sequences per round & 512 (SCP+MPPI's MPPI) & 512, or time-matched \\
Elites, weights & -- & best $J/8$; $\exp(\text{score}/0.5)$ (MPPI), uniform (CEM) \\
Sampling & -- & initial std 0.5 of the action range, variance in $[0.0025, 1]$ \\
Warm start & previous plan, shifted & previous mean, shifted; variance reset \\
\bottomrule
\end{tabular}
\caption{Planner settings.}
\label{tab:planners}
\end{table}

\subsection{Time matching}\label{app:timing}
We timed each planner on an idle GPU on every $N = 10$ model at $\epsilon = 0.2$ (3 episodes each, the first two steps of an episode excluded as compilation). SCP+MPPI took 0.40 to 0.71\,s per step on average across the six models. MPPI and CEM took 0.03\,s with 512 sequences per round, 0.33\,s with 32,768 and 1.43\,s with 131,072, on every model. The time-matched budget of 65,536 lies between these, at about 0.70\,s by linear interpolation, matching SCP+MPPI's mean on the slowest model and exceeding it on the other five. At $N = 20$ and 30 the budgets were matched to the plan times logged in the runs, 95,040 sequences took 2.04--2.05\,s against SCP+MPPI's 2.16\,s, and 120,000 took 3.81\,s against 3.61\,s. Each run used one 1g.10gb slice of an NVIDIA A100, with the QP solved on the CPU.

\FloatBarrier
\section{Statistical Comparisons}\label{app:stats}
Every comparison pairs an episode of SCP+MPPI with the other planner's episode on the same world model, margin and seed; the seed fixes the layout, the initial state and the planners' random numbers. For violations we report the difference in violation rate with Newcombe's score interval for paired proportions~\citep[method 10]{newcombe1998improved}, which stays accurate when violations are rare, and $P(\text{safer})$, the posterior probability that SCP+MPPI is the one that violates in fewer than half of the discordant pairs (pairs in which exactly one of the two planners violates), under a uniform prior on that share. For returns we report the mean paired difference with a 95\% bootstrap interval that resamples whole seeds ($10^4$ resamples), and $P(\text{higher return})$, the share of Bayesian-bootstrap draws~\citep{rubin1981bayesian}, flat Dirichlet weights over the seeds, in which the weighted mean difference is positive. A probability near 0.5 is no evidence either way, and $2(1 - P)$ approximates a two-sided $p$-value. Single planners' violation rates carry Wilson intervals and their mean returns bootstrap intervals over episodes.

Tables~\ref{tab:pooled_c1_2} and~\ref{tab:pooled_cinf} repeat Table~\ref{tab:baseline_comparison_N10} on the $c = 1.2$ and unbounded world models, on which SCP+MPPI violates slightly more often than the time-matched samplers (by 1.7 to 3.8 points), though none of these differences is significant. Pooled over all six $N = 10$ models (720 pairs), the differences from time-matched MPPI and CEM are $-0.1$ points [$-2.8$, $+2.5$] and $+0.7$ [$-2.2$, $+3.5$] in violation rate, and $+0.06$ [$-0.09$, $+0.20$] and $+0.09$ [$-0.05$, $+0.23$] in return.

\begin{table}[htbp]
  \centering\small
  \begin{tabular}{lrcccc}
\toprule
Baseline & Pairs & $\Delta$ violations (pp) & $P$(safer) & $\Delta$ return & $P$(higher return) \\
\midrule
SCP & 240 & $-23.3$ [$-30.8$, $-15.4$] & $>0.999$ & $+0.11$ [$-0.18$, $+0.41$] & 0.77 \\
MPPI & 240 & $+1.7$ [$-4.0$, $+7.3$] & 0.28 & $+0.07$ [$-0.21$, $+0.36$] & 0.68 \\
CEM & 240 & $+3.8$ [$-1.7$, $+9.2$] & 0.09 & $+0.11$ [$-0.22$, $+0.43$] & 0.76 \\
\bottomrule
\end{tabular}
  \caption{As Table~\ref{tab:baseline_comparison_N10}, on the two $c = 1.2$ models.}
  \label{tab:pooled_c1_2}
\end{table}

\begin{table}[htbp]
  \centering\small
  \begin{tabular}{lrcccc}
\toprule
Baseline & Pairs & $\Delta$ violations (pp) & $P$(safer) & $\Delta$ return & $P$(higher return) \\
\midrule
SCP & 240 & $-30.0$ [$-37.0$, $-22.5$] & $>0.999$ & $-0.17$ [$-0.49$, $+0.20$] & 0.15 \\
MPPI & 240 & $+1.7$ [$-3.8$, $+7.1$] & 0.28 & $+0.04$ [$-0.12$, $+0.20$] & 0.66 \\
CEM & 240 & $+1.7$ [$-3.6$, $+7.0$] & 0.27 & $-0.02$ [$-0.21$, $+0.18$] & 0.42 \\
\bottomrule
\end{tabular}
  \caption{As Table~\ref{tab:baseline_comparison_N10}, on the two unbounded ($c = \infty$) models.}
  \label{tab:pooled_cinf}
\end{table}

\FloatBarrier
\section{Additional Results}\label{app:results}

\subsection{Both sampling budgets}
\begin{figure}[htbp]
    \centering
    \includegraphics[width=0.85\linewidth]{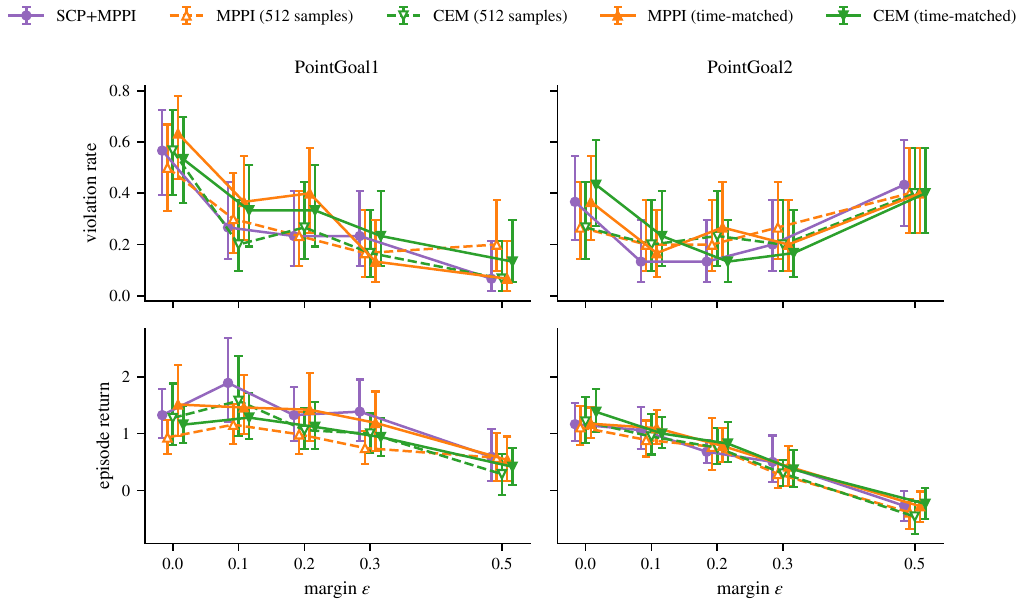}
    \caption{Fig.~\ref{fig:eps_sweep_1_5} with the samplers at both budgets: 512 sequences per round (open markers, dashed) and time-matched (filled, solid). Violation rate (top) and return (bottom) against the margin $\epsilon$ on each task's $c = 1.5$ model at $N = 10$, with 95\% intervals.}
    \label{fig:samplevtimematch}
\end{figure}

\begin{figure}[htbp]
    \centering
    \includegraphics[width=0.85\linewidth]{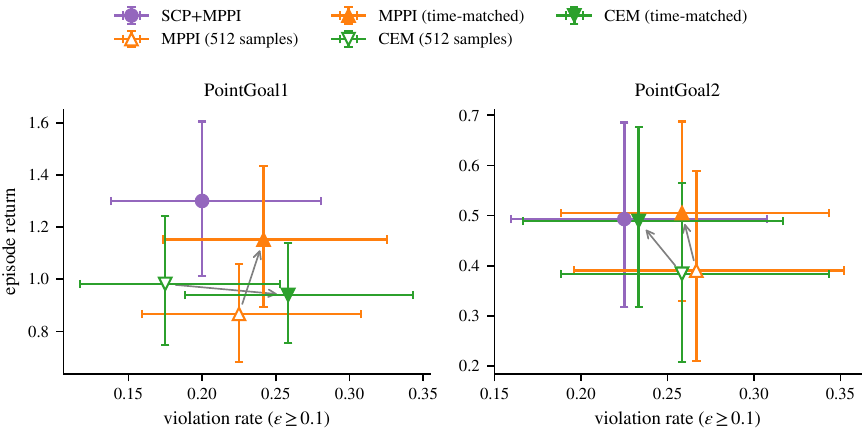}
    \caption{Return against violation rate on each task's $c = 1.5$ model at $N = 10$, pooled over $\epsilon \ge 0.1$, with 95\% intervals. Arrows lead from each sampler with 512 sequences per round to the same sampler time-matched.}
    \label{fig:returnvviol}
\end{figure}

Figs.~\ref{fig:samplevtimematch} and~\ref{fig:returnvviol} show what 128 times as many sequences buy the samplers on the $c = 1.5$ models. On PointGoal1 the extra sequences raise MPPI's return from 0.87 to 1.15 but not CEM's (0.98 to 0.94), and both violate more often with them (0.23 to 0.24 and 0.18 to 0.26), while SCP+MPPI earns 1.30 at 0.20. On PointGoal2 they raise the return (0.39 to 0.51 and 0.38 to 0.49) with little change in the violation rate.

\FloatBarrier
\subsection{Every world model}
\begin{figure}[htbp]
    \centering
    \includegraphics[width=\linewidth]{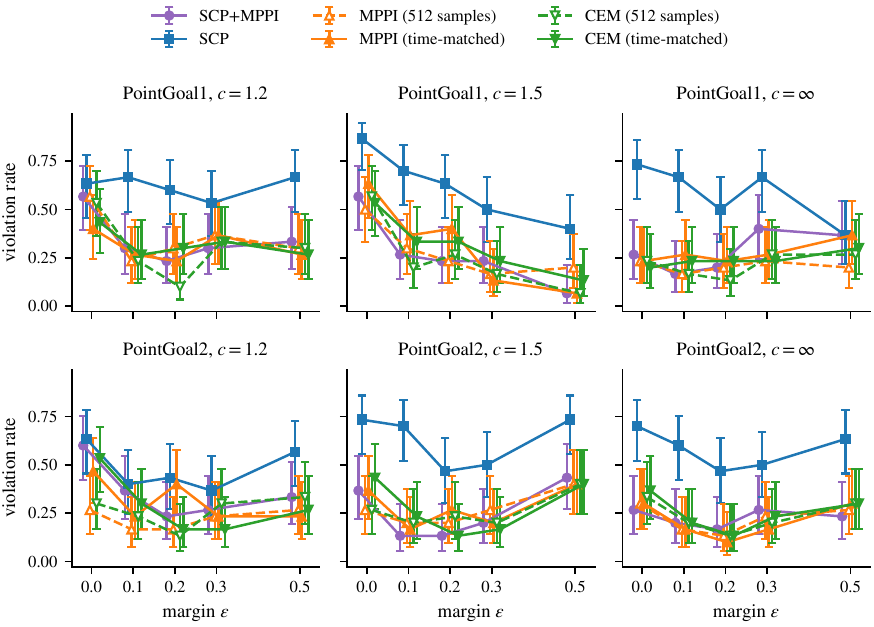}
    \caption{Violation rate against the margin $\epsilon$ for every planner on every $N = 10$ world model (rows: tasks; columns: Lipschitz bound $c$), with 95\% Wilson intervals.}
    \label{fig:eps_sweep_viol}
\end{figure}

\begin{figure}[htbp]
    \centering
    \includegraphics[width=\linewidth]{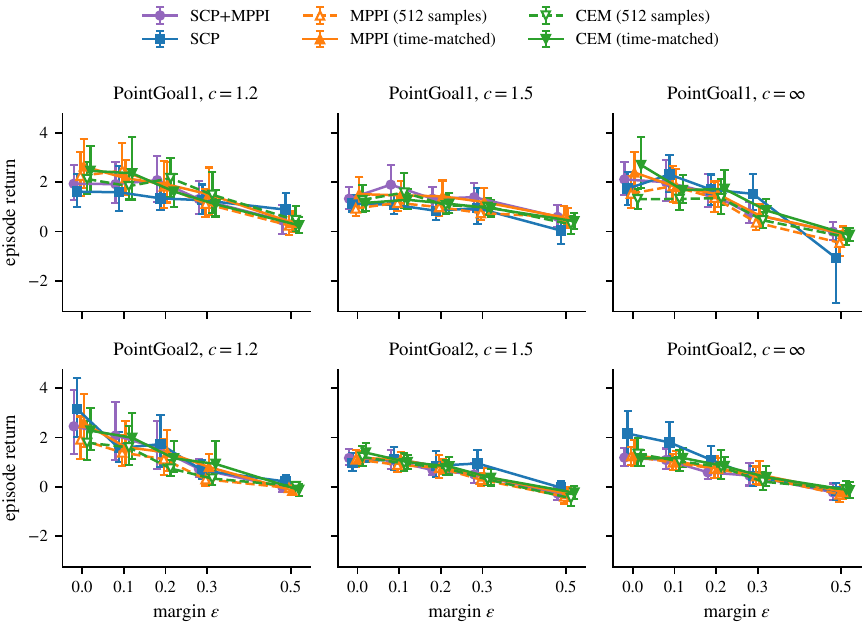}
    \caption{Return against the margin $\epsilon$ for every planner on every $N = 10$ world model (rows: tasks; columns: Lipschitz bound $c$), with 95\% bootstrap intervals.}
    \label{fig:eps_sweep_returns}
\end{figure}

Figs.~\ref{fig:eps_sweep_viol} and~\ref{fig:eps_sweep_returns} show the margin sweep on all six $N = 10$ models. Plain SCP is the least safe planner on every model, and on five of the six the violation rate of SCP+MPPI is lowest at $\epsilon = 0.1$ or 0.2.

\FloatBarrier
\subsection{Horizon}
\begin{figure}[htbp]
    \centering
    \includegraphics[width=0.85\linewidth]{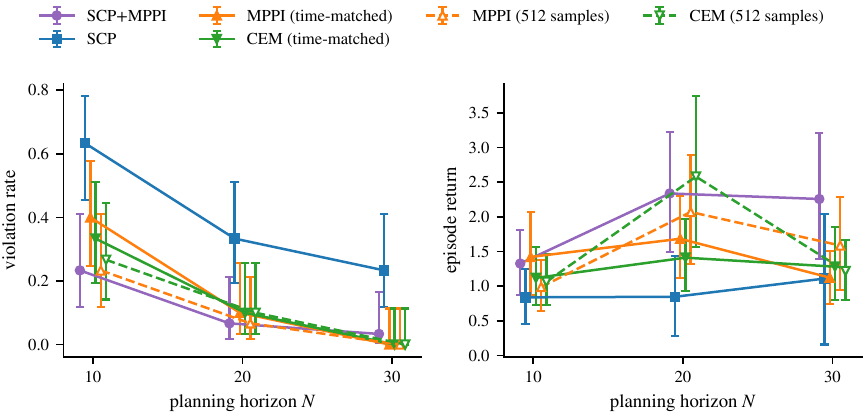}
    \caption{Violation rate (left) and return (right) against the horizon $N$ on PointGoal1 ($c = 1.5$, $\epsilon = 0.2$), with the samplers at 512 sequences per round (open markers, dashed) and at their time-matched budget (filled, solid), with 95\% intervals. Each $N$ uses its own world model, trained on windows of $N + 1$ steps, 30 episodes per point.}
    \label{fig:horizon_all_budgets}
\end{figure}

\begin{figure}[htbp]
    \centering
    \includegraphics[width=0.85\linewidth]{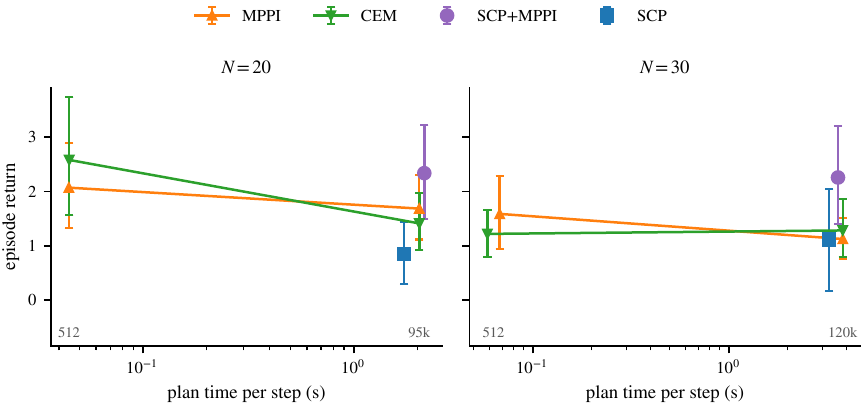}
    \caption{Return against mean plan time per step at $N = 20$ and 30, with 95\% intervals. Each sampler is shown at 512 sequences per round and at its time-matched budget (along the bottom), SCP+MPPI and SCP are single points.}
    \label{fig:return_compute}
\end{figure}

With 512 sequences per round, the samplers' return at $N = 20$ is close to SCP+MPPI's (MPPI 2.07, CEM 2.58, against 2.34; neither difference resolved), and at $N = 30$ below it (1.59 and 1.22 against 2.26; resolved for CEM;
Figs.~\ref{fig:horizon_all_budgets} and~\ref{fig:return_compute}).

\FloatBarrier
\subsection{The learned barrier as a monitor}
\begin{figure}[htbp]
    \centering
    \includegraphics[width=0.75\linewidth]{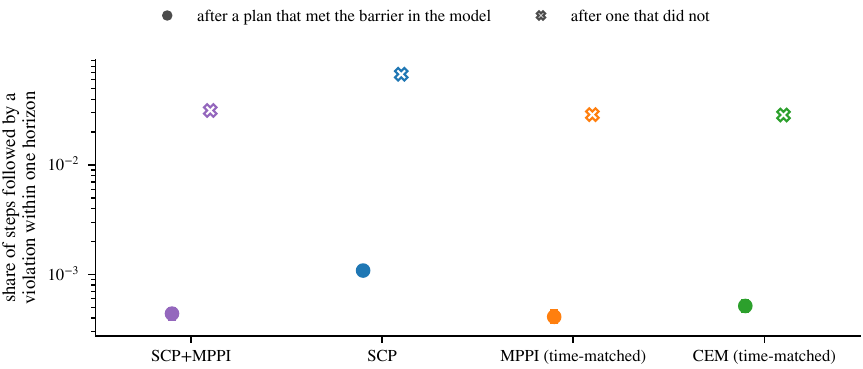}
    \caption{Share of steps followed by a violation within $N$ steps, after a plan that met the barrier in the model
    (dots) and after one that did not (crosses), pooled over the six $N = 10$ models and $\epsilon \ge 0.1$, with 95\%
    Wilson intervals (log scale).}
    \label{fig:latent_safe_plans}
\end{figure}

\begin{figure}[htbp]
    \centering
    \includegraphics[width=0.85\linewidth]{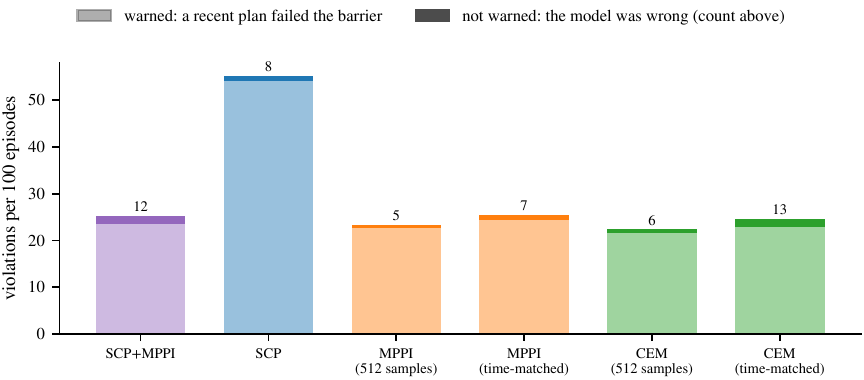}
    \caption{Violations per 100 episodes over the six $N = 10$ models and $\epsilon \ge 0.1$ (720 episodes per planner),
    split into those preceded within $N$ steps by a plan that failed the barrier in the model (warned, light) and those
    that were not (not warned, dark, counts above the bars).}
    \label{fig:crash_types}
\end{figure}

Fig.~\ref{fig:latent_safe_plans} gives the numbers behind Section~5.3, a violation follows within $N$ steps of a plan that met the barrier in the model about once in 2,000 steps for every planner but plain SCP (once in 900), and about once in 30 after a plan that did not (plain SCP: once in 15). Few violations come without such a warning (Fig.~\ref{fig:crash_types}): SCP+MPPI has 12 in
720 episodes, against 5 and 6 for the samplers with 512 sequences and 7 and 13 time-matched, paired by seed, none of these differences is resolved (McNemar $p \ge 0.12$).

\FloatBarrier
\subsection{When violations happen}
\begin{figure}[htbp]
    \centering
    \includegraphics[width=0.85\linewidth]{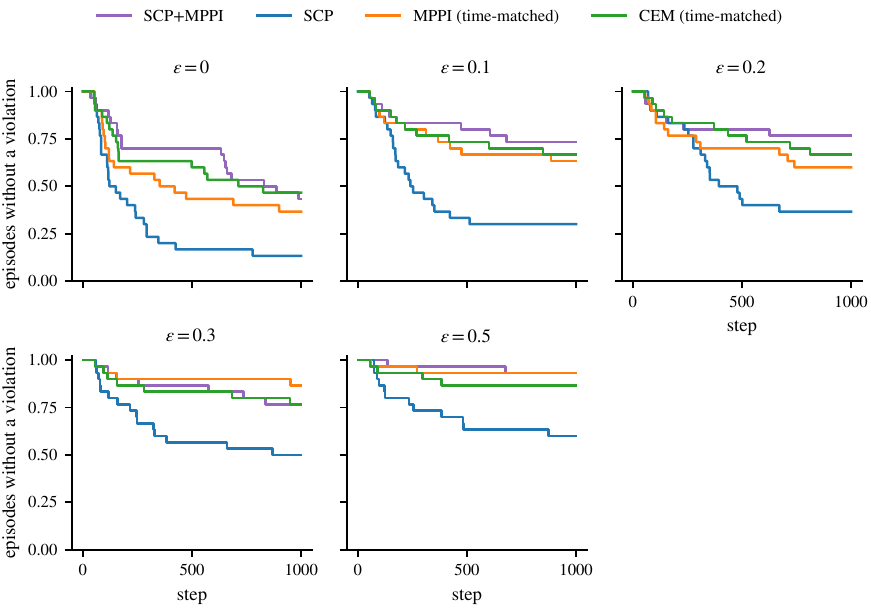}
    \caption{Share of episodes without a violation by each step, at each margin, on PointGoal1's $c = 1.5$ model at
    $N = 10$. Episodes end early only at a violation.}
    \label{fig:survival}
\end{figure}

At $\epsilon \ge 0.1$, about two thirds of the violations happen in the first 300 steps of an episode (Fig.~\ref{fig:survival}), plain SCP's come later (median step 234, against 154 to 179 for the other planners).

\FloatBarrier
\subsection{Model optimism}
\begin{figure}[htbp]
  \centering
  \includegraphics[width=\linewidth]{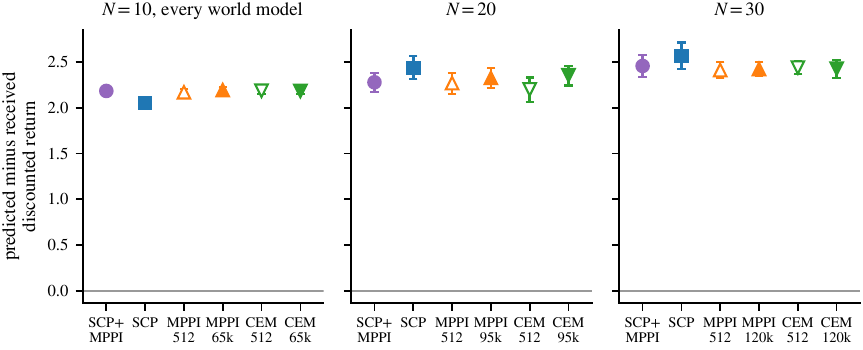}
  \caption{Model optimism: each executed plan's predicted return minus the discounted return received from that step
  on, averaged over each episode, with 95\% bootstrap intervals. $N = 10$ pools every world model
  ($\epsilon \ge 0.1$); $N = 20$ and 30 are the horizon study's models.}
  \label{fig:optimism}
\end{figure}

A gradient-based optimiser might exploit model error, finding plans whose predicted return the environment does not deliver. For each executed plan we compare its predicted return, the discounted predicted rewards plus the terminal value that carries the goal bonus, with the discounted return received from that step on, using steps at least 299 steps before the episode ends so that the unseen remainder weighs under 5\%. Every planner is optimistic by about 2, as expected of a value fitted to an upper expectile, but by nearly the same amount (Fig.~\ref{fig:optimism}): 2.05 (SCP) to 2.19 at $N = 10$, 2.20 to 2.44 at $N = 20$ and 2.41 to 2.56 at $N = 30$. SCP+MPPI is inside the samplers' range at
$N = 10$ and 20, and 0.02 above it at $N = 30$, where its paired differences from the samplers (+0.02 to +0.05) all
have intervals that include zero. Plain SCP is the most optimistic planner at $N = 20$ and 30, but we find no evidence
that SCP+MPPI exploits the model more than sampling does.

\end{document}